# Do we need to go Deep? Knowledge Tracing with Big Data

**Varun Mandalapu**[1], **Jiaqi Gong**[2], **Lujie Chen**[1]

[1]Department of Information Systems, University of Maryland Baltimore County, Baltimore, MD, 21250 USA
[2]Department of Computer Science, The University of Alabama, Tuscaloosa, AL, 35487 USA
varunm1@umbc.edu, jiaqi.gong@ua.edu, lujiec@umbc.edu

**Abstract**

Interactive Educational Systems (IES) enabled researchers to trace student knowledge in different skills and provide recommendations for a better learning path. To estimate the student knowledge and further predict their future performance, the interests in utilizing the student interaction data captured by IES to develop learner performance models is increasing rapidly. Moreover, with the advances in computing systems, the amount of data captured by these IES systems is also increasing that enables deep learning models to compete with traditional logistic models and Markov processes. However, it is still not empirically evident if these deep models outperform traditional models on the current scale of datasets with millions of student interactions. In this work, we adopt EdNet, the largest student interaction dataset publicly available in the education domain, to understand how accurately both deep and traditional models predict future student performances. Our work observes that logistic regression models with carefully engineered features outperformed deep models from the extensive experimentation. We follow this analysis with interpretation studies based on Locally Interpretable Model-agnostic Explanation (LIME) to understand the impact of various features on best performing model predictions.

## Introduction

Technological advances in interactive learning environments have come to force to provide personalized and high-quality education to students, especially accelerating learning and cutting teaching and training costs. This promise has successfully initiated the development process of the Interactive Educational Systems (IES) to teach different skills, such as English and mathematical problem solving. These systems integrate computational models based on 'learning by doing' principles to assess the students' knowledge and understanding and then provide relevant material and individual feedback. Students learning experience through these IES is as good as learning with a human tutor, which has been validated by previous research (VanLehn, 2011). However, one significant difference is that the cost of utilizing IES is low compared to a personal human tutor. Therefore, these IES systems are gaining prominence in the developing countries that lack enough qualified human tutors, such as Southeast Asian countries, compared to the countries in North America (Hao, 2019).

Modeling learner performance by tracing their interactions and knowledge over time is the primary characteristic of IES systems currently in production (Pelanek, 2017). This knowledge tracing and performance modeling also present some unique challenges in the Artificial Intelligence in Education (AIED) and Educational Data Mining (EDM) domains. Given the history of learner interaction in IES, the learner performance models estimate their current knowledge and support the prediction of future learner performances such as next problem correctness or the next solution for a similar skill or problem.

Advances in data science methodologies and the widespread availability of IES systems enabled researchers to develop data-driven models to track learner performances. This modeling of learner performances has three significant implications: generate actionable insights that support better learning, develop adaptive instruction policy, and estimate the knowledge of learners to support their learning activity (Rose et al., 2019). Most of adaptive learning and instruction technologies require a learner to master one skill and move to another(Ritter et al., 2016; Koedinger et al., 2013).Another trend is to provide students with a feedback system in the form of their skill progress bar that supports learners metacognitive abilities (Bull and Kay, 2010), fosters learners confidence in the system, and facilitates discussion between learners and tutors. Ultimately, it is essential to interpret models from the parameters fitted and outputs delivered. These model interpretations will allow researchers, engineers, and instructors to develop actionable insights and further develop current IES systems to incorporating more accurate modeling methodologies that contribute to learning sciences (Rose et al., 2019).

The implications of learner performance tracking models are at odds with each other. To develop adaptive learning environments, predicting learner performance with high accuracy matters most, whereas to develop actionable insights, it is crucial to focus more on interpretability than model accuracy. To develop trustworthy learning model interfaces, both accuracy and interpretability of learner performance models play a pivotal role Prior studies that developed learner performance tracking models focused on developing models that balance the tradeoff between accuracy and interpretability. Recent work by Gervet et al. (2020) discusses this issue in a detailed evaluation of deep learning and traditional model and recommends adopting black box models that provide higher accuracy in learner performance tracking for instructional policy and interpretable models to refine learner modeling in the education domain. Their work also discussed that with a larger dataset, deep models performed well compared to traditional models. Building off of the work done by Gervet et al. (2020), our work focuses on evaluating the performance of deep models compared to carefully engineered traditional models on a large dataset ever released in the education domain. As part of this work, we also focus on adopting Locally Interpretable Model Explanations (LIME) to understand the importance of different features on model predictions (Ribeiro, Singh, and Guestrin, 2016).

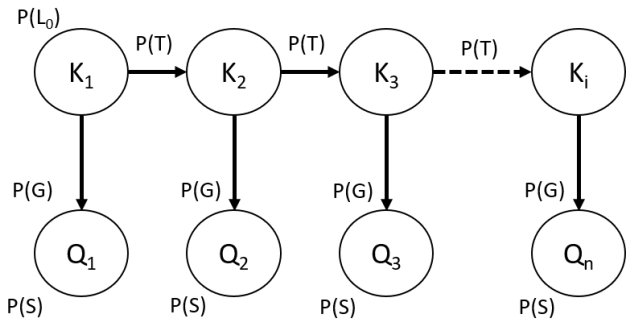

Figure 1: General representation of Knowledge tracing models

The development of deep models as an alternative to traditional statistical models like Bayesian Knowledge Tracing (BKT; Corbett and Anderson 1994) and Performance Factor Analysis (PFA; Pavlik et al. 2009) came into prominence. The general representation of knowledge tracing (KT) with question-solving logs is shown in below figure 1. The S represents the student node, the K values represent the corresponding Knowledge Components (KC) for each question, the Q represents the question accessed by a learner, the P(L) represents the initial learner knowledge, P(T) represents the knowledge gained by the learner that is transmitted to next question, P(G) and P(S) is the guess and slip probabilities for answering each question encountered by a learner in IES. In this work, we don't use BKT modeling as earlier studies (Gervet et al, 2020) showed that BKT models consistently performed worse than other learner performance models.

An earlier study that developed a deep model named Deep Knowledge Tracing (DKT; Piech et al. 2015) showed a gain of 25% in tracking learner performance compared to the traditional statistical model BKT on multiple real-world datasets. Later, a study performed by Xiong et al. (2016) showed that the initial performance increase showed by DKT over BKT is not substantial. An empirical study by Khajah et al. (2016) finds that BKT performs on par with DKT on four datasets with proper enhancements. In their paper, Gervet et al. (2020) evaluate multiple learner performance models by comparing deep models with traditional models on multiple datasets with varying sizes. Their work shows evidence that logistic regression (LR) models with carefully engineered features perform better on small and moderate-sized datasets, whereas deep models perform better on larger datasets or datasets that emphasize temporal information.

After carefully reviewing their results, we find that the deep model performance improvement is not much significant. These findings demonstrate the contributions of our work in two folds.

- Whether or not the datasets utilized in earlier studies are large enough to capitalize on the predictive power of deep learning models? This question is investigated by evaluating deep models on the largest dataset (EdNet; Choi et al, 2020) available in the education domain.
- To what extent the deep models perform better than traditional and interpretable models. This question is investigated by deploying traditional models, especially LR models, on carefully engineered features extracted based on different theories researched in the education domain.

In addition to these two questions, this work also studies the importance of different features extracted from EdNet data (Choi et al, 2020) on the best model that predicts future student performance. Our work adopts a correlation-based LIME method to explain different feature importance. Correlation-based LIME provides a correlation value between a feature and outcome variable for every sample in the dataset. Based on this correlation value, we categorize the variables into supporting or contradicting features.

The next sections in this paper review the earlier literature published in this area, discuss the developed approaches and methodologies to investigate the research questions, detail the dataset and evaluation metrics used, discuss the outcomes of this work in detail.

## Related Work

Learner modeling is at the forefront of developing highly effective IES. These models require continuous collection and updating of data related to learners in a well-defined mechanism. Developing a learner model consists of three important steps: gathering data of learner characteristics/interaction, constructing models based on the interaction, and updating the models by tracing learner activities on the IES. With the increased deployment of IES in educational settings, collecting a huge amount of data for modeling is accessible to researchers. The primary challenge is related to the development of accurate and interpretable models based on learner's activity and knowledge. A study by Chrysafiadi and Virvou (2013) focused on reviewing different modeling types based on different settings. This study showed that most of the researchers focused on combining multiple models to represent a wide variety of student learning characteristics. However, this review study focuses on learner modeling in general and not focused explicitly on IES. To understand various characteristics of developing learner models for IES, a study by Pelanek (2017) discusses the factors that influence choosing models for different learning contexts. This study also shows different issues in data collection, evaluation, and validation metrics.

As IES systems focus on accurate prediction of learner performance to improve adaptive learning, prior research focused on comparing different learner performance models to study their accuracy in predicting future learner performance. One extensive study done by Gervet et al. (2020) focused on comparing traditional and deep learning models' accuracy on different datasets related to ITS publicly available in the education domain. Their study shows evidence that deep learning models based on recurrent neural networks perform well on data set with a huge number of student interactions and has temporal dependencies. In contrast, traditional logistic regression models with features extracted based on Item Response Theory (IRT; van der Linden and Hambleton, 2013) and Performance Factor Analysis (PFA; Pavlik, Cen, and Koedinger, 2009) performed better on the datasets with a moderate number of samples and learners with multiple ITS interactions. Additionally, they also investigated the performance of the time window features studies in DAS3H (Choffin et al., 2019) on logistic regression models. This analysis showed that time window features did not add much predictive power to the logistic regression model compared to the deep models that capitalize on the dataset's temporal structures. However, one significant challenge is related to the impact of deep learning on big datasets. Compared to other domains like image recognition and sensor data, where deep learning performs very well, these models' impact in the education domain is not yet satisfactory. Therefore, an urgent need is to examine the intertwined relationship between modeling techniques and dataset characteristics.

## EdNet Dataset

EdNet is a large-scale dataset consisting of learner interaction with a multi-platform AI tutoring system named Santa (Choi et al, 2020). Santa facilitates tutoring English reading and listening to students interested in the Test of English for International Communication (TOEIC). It has 780,000 students from South Korea and available on iOS, Android, and Web. The systems adapt to learner inputs by providing them with relevant video lectures, expert commentaries, and assessing their solutions.

EdNet is the largest real-world IES student interaction dataset ever made available to the public in the education domain (Choi et al, 2020). It consists of more than 131 Million interactions collected from 784,309 students in the span of two years from 2017. This dataset enables researchers to solve some critical challenges in education with a specific focus on accurately predicting learner performance with IES. This dataset consists of 13,169 questions that were tagged to 293 skills and 1021 lectures. Each of them was consumed more than 95 Million times and 601,805 times, respectively. EdNet was organized by following a hierarchical structure. Each level of EdNet data has different data points and was named KT1, KT2, KT3, and KT4. As the hierarchy and postfix increases, the number of actions and the type of actions performed by learners in the dataset increases. As this work focused on comparing the predictive power of traditional and deep models in learner performance tracking, we adopt the KT1 dataset, the simplest form of all four datasets released by EdNet. KT1 is the only dataset consisting of interactions from 784,309 students, whereas KT2, KT3, and KT4 consist of interactions from fewer students, around 297,000.

KT1 is the simplest form of all the datasets provided in EdNet. It consists of student question logs that is the basic form of input features provided to different deep models like DKT and Self Attentive Knowledge Tracing (SAKT; Pandey and Karypis 2019). This dataset consists of five features:

- The *timestamp* is the time in milliseconds when the learner encounters a question.
- *Question_id* provides a unique ID for each question and represents by q{integer}.
- *Bundle_id* provides the unique bundle ID related to each question answered by the learner.
- *User_answer* is the feature that captures the multiple-choice answer between alphabets a to d chosen by the user for a particular question.
- *Elapsed time* is the amount of time in milliseconds spent by the learner on a particular question.

Some important statistics related to EdNet KT1 data are provided in the below table 1. These statistics were extracted after assigning KCs to questions and duplicating question samples if multiple KCs were assigned to each question. In this work, we use KC and skill tag alternatively as they both mean the same.

Table 1: The description of EdNet KT1 dataset

| Description | EdNet KT1 |
|---|---|
| Number of Students | 784,309 |
| Number of Interactions after expanding KCs | 224,461,772 |
| Number of KCs | 188 |
| Number of unique questions | 12,284 |
| Number of correct answers | 152,561,335 |
| Number of wrong answers | 71,900,437 |
| Collection Period | 2 Yr 7 Months |

## Methodology

The methodology section in this work is divided into three subsections. In the first subsection, we discuss the preprocessing steps taken and the characteristics of the EdNet KT1 dataset. In the second subsection, we discuss the features extracted from the EdNet KT1 dataset used in traditional models, and in the third subsection, we discuss the models used and detail the methodology used for explaining model predictions.

### Data Preprocessing

The original EdNet dataset consists of five features representing learner question-solving logs captured by the Santa tutoring platform. These five features in the original form need some preprocessing to fit into deep models. To do that, we first map the user noted answer with the correct answer of the original question and generate a flag that specified whether the user-provided answer is correct or wrong. This answer flag will act as the ground truth label to fit a learner performance model. To be consistent with prior work that focused on evaluating learner performance models on different datasets, our work adopts the same data preprocessing steps detailed by Gervet et al (2020). We first assign the KC tags to each question answered by the learner. In the next step, learners with fewer than ten interactions and learner interactions with unspecified or NaN KC tags were removed from the dataset. As each question can be tagged to multiple KCs, our work converts the unique combination of KCs into a new KC tag that will be used for DKT and SAKT algorithm. The statistics of the dataset after preprocessing were listed in Table 2.

Table 2: The description of EdNet KT1 dataset after preprocessing

| Description | EdNet KT1 |
|---|---|
| Number of Students | 607,610 |
| Number of Interactions | 93,193,461 |
| Mean KCs per item/question | 2.26 |
| Median items per KC | 48.5 |
| Median learners per item | 4201 |
| Median learners per KC | 118892 |
| Median interactions per learner | 21 |

It includes the total number of students present in the dataset after preprocessing, number of interactions, mean KC components per item/question, median items per KC, Median learner per item and median learners per KC We found there are a total of 1302 unique KC combinations in addition to 188 original KCs. Our work used original 188 KC tags for this analysis. Also, from table 2, it is evident that the total number of questions answered correctly was more than the questions answered incorrectly by all students.

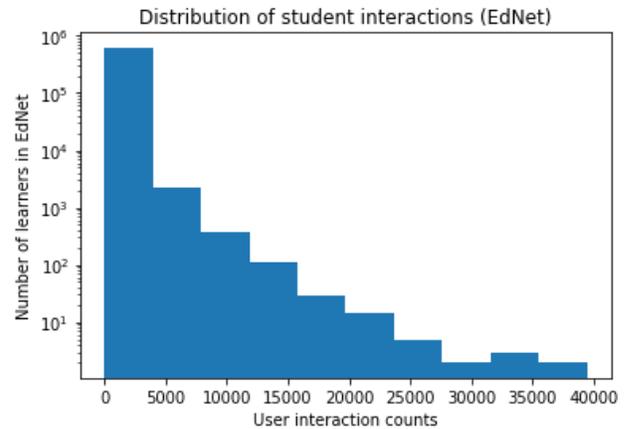

Figure 2: Power-law distribution exhibited by EdNet data for number of interactions per learner.

An earlier study that reviewed multiple student interaction datasets found that the dataset size directly influences learner prediction models' performance (Gervet et al, 2020). This phenomenon is based on the capability of different models to generalize on different sizes of datasets. Additionally, this work also shows that the number of parameters fitted linearly scales with the number of KCs and interactions. Based on these insights, the dataset size is also determined by the number of learners who attempt each KC and item. These statistics were available in Table 2 for all students in EdNet. Another important characteristic is the number of interactions per learner that influences the learner model performance. An earlier study showed that most datasets exhibit power-law distribution for the number of interactions recorded for each learner. To study the distribution of interactions per learner, we plotted the number of interactions vs. the number of learners, as shown in figure 2. Based on this plot, it is evident that the interaction per learner in EdNet data also follows the power-law distribution.

## Feature Extraction for Learner Performance Modeling

Learner performance prediction modeling can be formalized as a supervised sequence of learner tasks. Given the history of learner past interactions with the IES, learner performance models predict future learners characteristics. The EdNet dataset also specifies the KC for each question and extracts the learner's time on each question. These timestamps provide insights about the learner's time continuity of the problem, and the time interval between different actions aids in understanding learner practice patterns and forgetfulness outside of the IES. When tagged with learner's questions, knowledge components can be used to decompose the individual knowledge types targeted in the instruction. Some examples of knowledge components in English grammar can be nouns and verbs. The experts in the field will assign these knowledge components per question. It is also possible to assign multiple KCs for the same question as the question might be addressing different skills.

Learner performance models probability to predict learners correct answer flag depends on the learners master of a particular KC they are encountering. This mastery of skill can be extracted based on the learner's previous interactions and answers for items and skill types. This understanding of KCs will help distinguish the features that need to be extracted from raw IES interaction data for a specific algorithm. Two basic approaches were followed to develop learner performance models in supervised learner tasks. The first one is to extract features from the historical learner interactions with IES and the second one is to adapt algorithms that were designed to process variable-length data captured by the IES. The performance of algorithms in the first approach depends on the quality of features engineered from the historical learner interaction data, whereas the second approach depends on the ability of an algorithm to remember the past interactions for different KCs. To be consistent with earlier work in this domain, we extract the following features from EdNet dataset.

**Item Response Theory Feature Vector:** Item response theory (van der Linden and Hambleton 2013) is one of the most studied psychometric models that is also gaining popularity in computer-based adaptive testing environments and intelligent learning community. This model works on the principle that all questions given in a test are determined by a single skill referred to as a latent trait. In this work, we one hot encode the items into independent features that represent the learner's question. IRT predictions will take the logistic form as shown in below equation. The *C* value represent the answer correctness for a given question *Q*. The *a* value represents the ability of a learner and the *β* represents the difficulty of a given question. The models identify it from the one hot encoded feature extracted from item id. For complete details, please go through the referenced articles. To be consistent with earlier work (Gervet et al., 2020), we only one hot encode items/questions in the dataset and does not fit student ability into IRT based modeling.

$$p\left(\frac{C}{Q}\right) = \frac{1}{1+e^{(a-\beta_{qs})}}$$

**Performance Factor Analysis (PFA):** Performance factor analysis (Pavlik et al. 2009) is a modification of learning factor analysis. In PFA, the model takes student practice and on relevant skills into consideration. This consideration is due to two reasons. First, the previous correct answers for a particular skill define students' strength in that skill and strengthen the model estimates. Another reason is that correct response increases student learning compared to incorrect responses due to greater processing during correct responses. Please refer to Pavlik et al (2009). for more details on PFA. In line with the previous study (Gervet et al, 2020), our work extracts two important feature sets in PFA. The PFA features in this study consists of skill one hot encoding, rescaled value of past attempts for relevant skill and past wins (correct answers) for a relevant skill.

**DAS3H:** DAS3H is a student learning and forgetting model introduced by Choffin et al (2019). It stands for item/question difficulty, ability, skill, and skill practice history. Earlier model named DASH outperformed hierarchical Bayesian based IRT, but not scalable to multiple skill item tagging. Building off this earlier work, the DAS3H model was developed to scale for items with multiple KCs and consider the impact of the historical practice on current learner performance to differ multiple skills. As DAS3H is based on time windowing of data samples, our work also extracts all the mentioned features based on time windowing followed in DAS3H. For more details on DAS3H, please refer to the following article authored by Choffin et al. (2019). In this work, we extract the following features as a part of DAS3H: item one hot encoding, skills one hot encoding, count of past attempt per skill with time windowing and count of past wins per skill with time windowing.

**Best LR-Features (Gervet et al. (2020)):** Based on prior study by that examined multiple features sets, they observed that DAS3H without time windowing and augmenting with total count features performed better compared to all other feature sets discussed above. We also extract these features to test the performance of models. These features include: All DAS3H features without time windows, past attempts for item, past attempt for all items, past wins for item, past wins for all items attempts.

In addition to these features we also test the Best LR-features in Gervet et al. (2020) by including time windowing property. All the features sets are referred in Table 3 of results section.

**Algorithms & LIME**

In this study, we focus on three major algorithms to compare traditional models vs. deep models. First, we develop a *Baseline Model*. This baseline model will extract the correctness probability based on each item/question in the training set and will be applied as prediction to the test set data. This is similar to IRT, but without fitting any model. As part of the traditional models, we evaluate logistic regression, a popular algorithm in the education domain, to evaluate features extracted based on IRT, PFA, DAS3H, Best-LR features (Gervet et al. 2020) and Bes-LR features with time windowing. In deep models, we evaluate Deep Knowledge Tracing (DKT) and Self Attentive Knowledge Tracing (SAKT). As deep models discard handcrafted features in favor of features directly learner from data, we use the original dataset to feed into these models

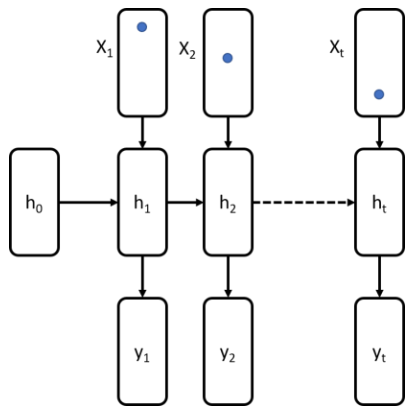

Figure 3: RNN representation of Deep Knowledge Tracing algorithm. The hidden states are represented by the tanh function and the input features are one hot encodings of user items.

DKT is a Recurrent Neural Network (RNN) based algorithm that gains information from the temporal structures in the data. Figure 3 below shows the RNN representation of the DKT algorithm. The input features $X_1$ to $X_t$ are the unique one-hot encoding of interactions at a given time step. These encoded inputs were fed into the hidden nodes labeled from $h_1$ to $h_t$. These nodes are represented as the successive summary of information from historical observations related to future predictions. The nodes will then be connected to an output vector $y_1$ to $y_t$ representing the probability of predicting learner answers correctly.

Multiple other deep models were developed post DKT introduction, but they showed little improvement than the original DKT model, with one notable exception of SAKT. SAKT works based on an attention mechanism that first visits the past interactions of a learner to predict future performances. In this process, SAKT first computes the similarity between question embeddings and past interaction encodings. A convex combination of past samples was weighed by normalized similarity score, and finally, the convex combination is transformed into linear form and transmitted through a sigmoid function to obtain the probability of future predictions.

**LIME for Explanations**: One crucial area we explored in this work is understanding the importance of input features on algorithm predictions. To do this, we adopted an interpretation method based on Locally Interpretable Model-agnostic Explanations (LIME), as shown in figure 4. In this method, we first select the algorithm that performs better in predicting test data of EdNet KT1. Once the algorithm is chosen, we generate new samples for each sample present in the test data and assign labels based on the model predictions. Finally, we calculate the correlation weights between features and the outcome variable. Once the correlation weights for features are available for every sample, we divide the samples based on correct and incorrect predictions. Doing this will help us understand the features that contributed and contradicted both correct and incorrect features (Mandalapu and Gong, 2019). This method of implementing LIME is different from the original LIME proposed by Ribeiro et al. (2016), as the original method is slow and not compatible with categorical features.

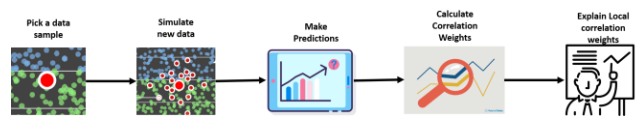

Figure 4: Correlation base Locally Interpretable Model Explanation process.

**Experimental Setup:** This study adopted the python platform for data preprocessing, feature extraction, and model evaluation. To be consistent with earlier work, we modified the codes provided by Gervet et al. (2020) to suit EdNet dataset. The logistic regression model with L-BFGS solver was adopted from the Scikit-learn package (Pedregosa et al., 2011). The default hyperparameters were kept in evaluating models as earlier works observed no improvement with hyperparameter optimization on ITS datasets. The deep learning models DKT and SAKT were implemented in PyTorch

(Paszke et al., 2017). For DKT hyperparameters, we adopted Adam optimizer, a learning rate of 0.001, dropout in the range of 0 to 0.5 in 0.25 increment, the batch size is adjusted based on the GPU size, and a hidden state dimension of 200. The SAKT sequence length is decided based on the median number of interactions per learner. For SAKT, we also selected a single attention layer with a drop probability of 0.25 and an embedding dimension of 200. The LIME model explanations are implemented in the RapidMiner (Mierswa et al. 2006) data science tool.

One challenge we encountered with the dataset is related to the size and time complexity. For logistic regression model features extracted using IRT, PFA & DAS3H, the dataset size is very high, and the extraction time is not feasible for implementation with existing resources. Earlier research on multiple ITS datasets showed that DKT over fitted for small datasets and logistic regression model overfitted for very large datasets. As a workaround, we randomly selected 50000 students from the EdNet dataset and extracted features for all model evaluations. The skills, items, and correctness flag of these 50000 students are in proportion with the overall dataset. We performed a split validation where the dataset is divided into 80:20 percent split based on learner population. So, all the learner samples are present either in training or in the testing dataset but not both. Our study compares all the models based on the Area Under Curve (AUC) metric that is popular in education research.

## Results

In this section, we detail the results of different models and LIME based feature importance's. Table 3 below shows the AUC values on test data for both logistic regression (LR) models and the deep learning models on 50000 students dataset. Table 4 compares deep models on all EdNet KT1 data related to 607,610 students. For feature set details, please refer to *Feature Extraction for Learner Performance Modeling* subsection in *Methodology* section of this paper.

From both tables 3 and 4, it is evident that a logistic regression model evaluated on all Best LR-Features (Gervet et al.) with Time windowing performed better compared to any other model evaluated in this study. This finding is consistent with an earlier study (Gervet et al, 2020) that compared traditional and deep models on different ITS datasets. One interesting observation is that DKT's performance reduced when trained on more samples compared to logistic regression. This outcome is in contradiction with another study (Gervet et al, 2020) that showed DKT performs slightly better in large datasets. SAKT model showed similar performance when trained on both 50000 and 607,610 students dataset. In an earlier study (Gervet et al, 2020), the authors observed that SAKT underperformed compared to DKT on the ASSISTment 2015 dataset. This is the largest dataset analyzed in that study. Authors in that study hypothesized that ASSISTment 2015 is small for SAKT (Gervet et al, 2020). This hypothesis seems true as we observe that SAKT outperformed DKT on the largest dataset (EdNet) ever released in the education domain.

Table 3: EdNet dataset (50000 students) performance for all models evaluated in this study

| Feature Set | Model | AUC |
|---|---|---|
| Correctness Probability per Item | Baseline Model | 0.73 |
| IRT | LR | 0.73 |
| PFA | LR | 0.66 |
| DAS3H | LR | 0.76 |
| Best LR-Features (Gervet et al. 2020) | LR | 0.68 |
| Best LR-Features with Time windowing | LR | 0.77 |
| Original features | DKT | 0.76 |
| Original features | SAKT | 0.76 |

Table 4: EdNet dataset (607,610 students) performance for all models evaluated in this study

| Feature Set | Model | AUC |
|---|---|---|
| Original Features | DKT | 0.72 |
| Original Features | SAKT | 0.76 |

Table 5: LIME based feature importance's for logistic regression model trained on Best-LR features with time windows.

| Features | Correct Predictions | | Incorrect Predictions | |
|---|---|---|---|---|
| | Support | Contradict | Support | Contradict |
| KCs (One hot encoded) | 0.026 | -0.028 | 0.014 | -0.014 |
| Attempt (Counts) | 0.023 | -0.028 | 0.012 | -0.014 |
| Wins (Counts) | 0.021 | -0.026 | 0.011 | -0.013 |
| Item (One hot encoded) | 0.028 | -0.028 | 0.014 | -0.014 |

In our work, we also explored the importance of features based on LIME (Mandalapu and Gong, 2019). As the best algorithm is a logistic regression model trained on Best-LR features with time windowing, we evaluated it with LIME on 1000 students randomly selected from the test dataset to reduce time and space complexity. As LIME generates new samples (300 in this study) for every learner interaction in the dataset, even with 1000 students, it requires a huge amount of memory and time. Once LIME is applied, we extract feature importance metrics for correct and incorrect predictions, as shown in table 5. The best features are decided based on the tradeoff between the importance of supporting attribute in correct predictions and importance of contradicting attribute in incorrect predictions.

Table 6: Importance of difficult and easy skills in model predictions.

| Top 3 Difficult Skills | Correctness Ratio | LIME Importance | Top 3 Easy Skills | Correctness Ratio | LIME Importance |
|---|---|---|---|---|---|
| 142 | 0.38 | -0.038 | 3 | 0.81 | 0.032 |
| 145 | 0.43 | 0.004 | 9 | 0.85 | 0.05 |
| 169 | 0.43 | 0.009 | 16 | 0.86 | 0.0014 |

Based on the LIME explanations, we observe that the feature set related to item one hot encoding supported the model in predicting the correct label. This observation is also supported by the model evaluation metric reported in table 3. A logistic regression model trained on IRT features has an AUC of 0.73. The feature set corresponding to wins (correct answers) in student question logs has less impact on model predictions. This phenomenon can be observed in the performance of PFA (AUC = 0.66) as in the absence of item related features, the attempt and win related features perform below any other model trained in this study.

LIME helps understand the relationship between "toughness of skill" and its importance in the model predictions. As the dataset is anonymized, we hypothesize that the "toughness of skill" can be determined based on the correctness ratio. The correctness ratio is the ratio of correct answers per skill by the total number of interactions per skill. Table 6 shows the top 3 skills that are easy to master by the learner and the top 3 skills that are difficult to master based on the correctness ratio and their respective local importance. The positive value of importance indicates particular skill support correct predictions, and the negative value indicates a particular skill contradicts correct predictions. From Table 6, we can observe that the hardest skill is contradicting correct predictions. This might be due to inconsistency in learner answers as it is hard to master a skill with high difficulty based on the EdNet dataset. One caveat: the importance shown in Table 6 are independent local feature importance's. Model prediction performance is also dependent on complex interactions between features at a global level.

## Discussion

This work evaluates different learner performance models on the largest student interactions dataset (EdNet) released in the education domain. The results show that logistic regression models with carefully handcrafted features based on different education theories and expert tags perform better than deep learning models that are complex and hard to interpret. One caveat: the logistic regression model is evaluated on 50000 students randomly selected from the total dataset available in EdNet KT1. This selection of a limited number of students for the logistic regression model is due to the space complexity as handcrafted features for 607K students take a tremendous amount of space and the amount of extraction time needed is not feasible in real-world applications. Earlier studies found that deep learning models like DKT and SAKT scaled well for datasets with larger student interactions and temporal structures, but they also hypothesize that the largest dataset (ASSISTment 2015) utilized in those studies is still not sufficient for deep models' convergence. Our work also evaluates this hypothesis by evaluating deep models on 93 Million interactions generated by 607K students in the Santa tutoring platform. The results conclude that the deep model's test performance is similar to the logistic model on 50000 students, and with all student's data, DKT has a reduction in performance. This might be due to the difference in interactions per learner between a dataset with 50000 students and 607K students. Earlier study (Gervet et al. 2020) also noted that datasets with high number of interactions per learner have a reduction in DKT performance over logistic regression model. This is also due to inability of DKT to keep track of long-term information as reported by Hochreiter et al., (2001).

In addition to evaluating multiple learner performance models, this work also focuses on interpreting feature sets that played an important role in the best performing model predictions. We explore the features based on a correlation LIME method that details feature correlation with predictions at a local level. This understanding of features at a local level will help researchers design systems that make highly accurate predictions of future learner performance.

Future work will be to explore learner heterogeneity, for example, by modeling explicability of the ability in IRT. Further analysis of IRT should also focus on the amount of data needed for competitive modeling and representativeness. Another research topic could be studying the adaptive algorithm's effect on knowledge tracing and modeling student learning curve related to a skill. Based on the importance of IRT features in model predictions, it is important to study its scalability. One interesting question would be to identify the amount of item response data needed to reach satisfactory performance. This finding will help update trained models when a new question is added to the existing question bank.